\documentclass[10pt,twocolumn,letterpaper]{article}

\usepackage{cvpr}              

\usepackage{graphicx}
\usepackage{amsmath}
\usepackage{amssymb}
\usepackage{booktabs}
\usepackage{color}
\usepackage{colortbl}
\usepackage{xcolor}
\usepackage{bbm}
\usepackage{graphicx}
\usepackage{multirow}
\usepackage[title]{appendix}

\usepackage[pagebackref,breaklinks,colorlinks]{hyperref}
\usepackage[accsupp]{axessibility}  

\usepackage[capitalize]{cleveref}
\crefname{section}{Sec.}{Secs.}
\Crefname{section}{Section}{Sections}
\Crefname{table}{Table}{Tables}
\crefname{table}{Tab.}{Tabs.}

\definecolor{mygray}{gray}{.9}

\def\cvprPaperID{1786} 
\def\confName{CVPR}
\def\confYear{2023}

\begin{document}

\title{Anchor3DLane: Learning to Regress 3D Anchors for Monocular 3D \\Lane Detection}

\author{Shaofei Huang\textsuperscript{\rm 1,2} \quad Zhenwei Shen\textsuperscript{\rm 3}\thanks{Work done while at TuSimple} \quad Zehao Huang\textsuperscript{\rm 3} \quad Zi-han Ding\textsuperscript{\rm 4,5} \\ \quad Jiao Dai\textsuperscript{\rm 1,2} 
\quad Jizhong Han\textsuperscript{\rm 1,2} \quad Naiyan Wang\textsuperscript{\rm 3} \quad Si Liu\textsuperscript{\rm 4,5} \\
\textsuperscript{\rm 1} Institute of Information Engineering, Chinese Academy of Sciences\\
\textsuperscript{\rm 2} School of Cyber Security, University of Chinese Academy of Sciences\\
\textsuperscript{\rm 3} TuSimple \quad 
\textsuperscript{\rm 4} Institute of Artificial Intelligence, Beihang University \\
\textsuperscript{\rm 5} Hangzhou Innovation Institute, Beihang University \\
{\tt\small \{nowherespyfly, zehaohuang18, zihanding819, winsty\}@gmail.com} \\
{\tt\small shenzhenwei@outlook.com \quad \{hanjizhong, daijiao\}@iie.ac.cn \quad liusi@buaa.edu.cn}
}

\maketitle

\begin{abstract}
Monocular 3D lane detection is a challenging task due to its lack of depth information.
A popular solution is to first transform the front-viewed (FV) images or features into the bird-eye-view (BEV) space with inverse perspective mapping (IPM) and detect lanes from BEV features.
However, the reliance of IPM on flat ground assumption and loss of context information make it inaccurate to restore 3D information from BEV representations. 
An attempt has been made to get rid of BEV and predict 3D lanes from FV representations directly, while it still underperforms other BEV-based methods given its lack of structured representation for 3D lanes.
In this paper, we define 3D lane anchors in the 3D space and propose a BEV-free method named Anchor3DLane to predict 3D lanes directly from FV representations.
3D lane anchors are projected to the FV features to extract their features which contain both good structural and context information to make accurate predictions.
In addition, we also develop a global optimization method that makes use of the equal-width property between lanes to reduce the lateral error of predictions.
Extensive experiments on three popular 3D lane detection benchmarks show that our Anchor3DLane outperforms previous BEV-based methods and achieves state-of-the-art performances. 
The code is available at: \url{https://github.com/tusen-ai/Anchor3DLane}.
\vspace{-0.2cm}

\end{abstract}

\section{Introduction}
\label{sec:intro}

\begin{figure}[!t]
    \centering
    \includegraphics[width=\linewidth]{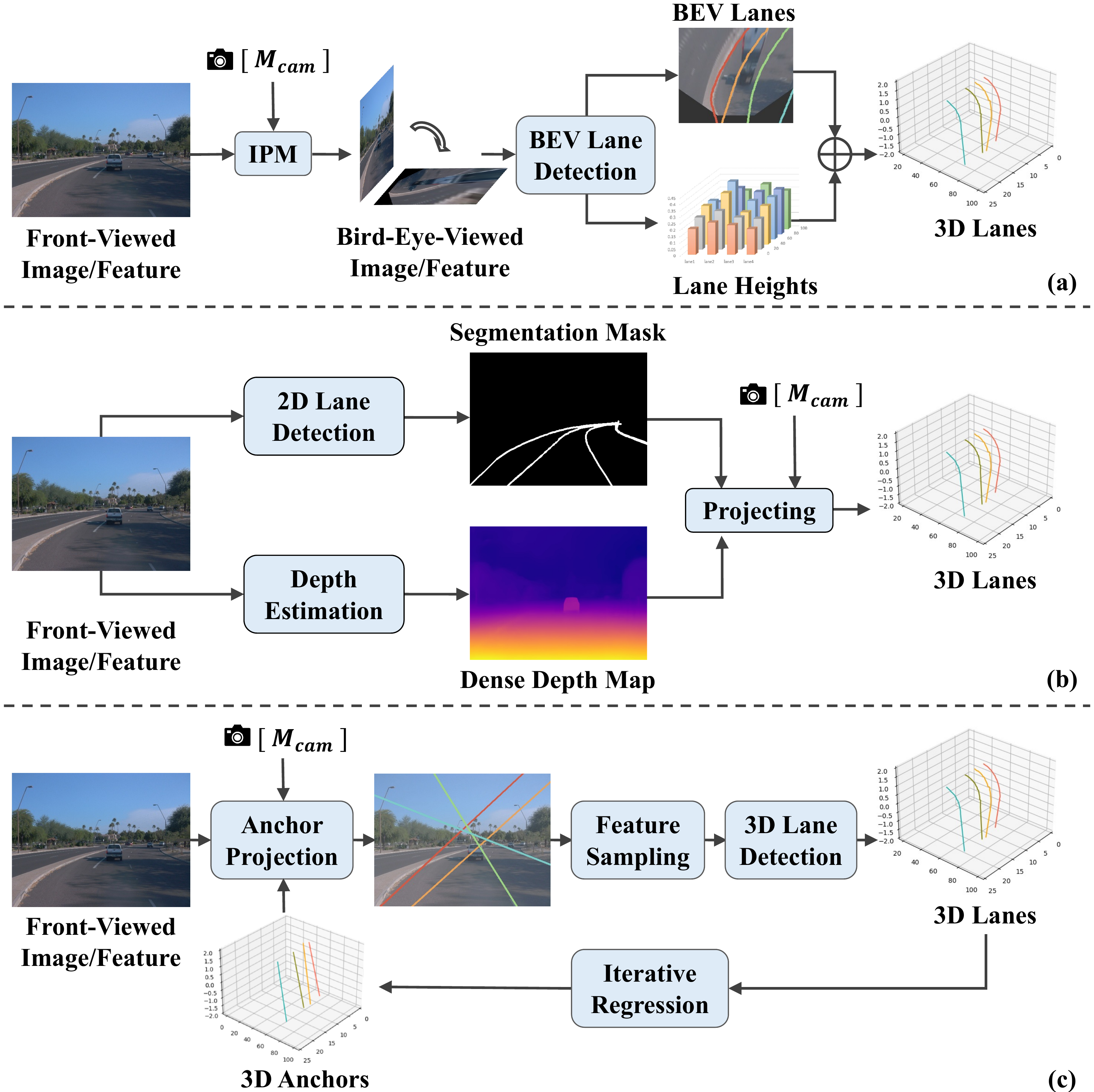}
    \caption{(a) BEV-based methods, which perform lane detection in the warped BEV images or features. (b) Non-BEV method, which projects 2D lane predictions back to 3D space with estimated depth. (c) Our Anchor3DLane projects 3D anchors into FV features to sample features for 3D prediction directly.}
    \vspace{-0.8cm}
    \label{fig:intro}
\end{figure}
Monocular 3D lane detection, which aims at estimating the 3D coordinates of lane lines from a frontal-viewed image, is one of the essential modules in autonomous driving systems. 
Accurate and robust perception of 3D lanes is not only critical for stable lane keeping, but also serves as an important component for downstream tasks like high-definition map construction~\cite{liu2020high, qin2020avp}, and trajectory planning~\cite{zhu2020trajectory, altche2017lstm}.
However, due to the lack of depth information, estimating lanes in 3D space directly from 2D image domain still remains very challenging.

A straightforward way to tackle the above challenges is to detect lanes from the bird-eye-viewed (BEV) space.
As illustrated in Figure~\ref{fig:intro}(a), a common practice of BEV-based methods~\cite{3dlanenet, genlanenet, persformer, clgo} is to warp images or features from frontal-viewed (FV) space to BEV with inverse perspective mapping (IPM), thereby transforming the 3D lane detection task into 2D lane detection task in BEV.
To project the detected BEV lanes back into 3D space, coordinates of the lane points are then combined with their corresponding height values which are estimated by a height estimation head.
Though proven effective, their limitations are still obvious:
(1) IPM relies on a strict assumption of flat ground, which does not hold true for uphill or downhill cases.
(2) Since IPM warps the images on the basis of ground, some useful height information as well as the context information above the road surface are lost inevitably.
For example, objects like vehicles on the road are severely distorted after warping.
Therefore, information lost brought by IPM hinders the accurate restoration of 3D information from BEV representations.

Given the above limitations of BEV, some works tried to predict 3D lanes from FV directly.
As illustrated in Figure~\ref{fig:intro}(b), SALAD~\cite{once} decomposes 3D lane detection task into 2D lane segmentation and dense depth estimation.
The segmented 2D lanes are projected into 3D space with camera intrinsic parameters and the estimated depth information.
Even though getting rid of the flat ground assumption, SALAD lacks structured representations of 3D lanes.
As a result, it is unnatural to extend it to more complex 3D lane settings like multi-view or multi-frame. Moreover, their performance is still far behind the state-of-the-art methods due to the unstructured representation.

In this paper, we propose a novel BEV-free method named Anchor3DLane to predict 3D lanes directly from FV concisely and effectively.
As shown in Figure~\ref{fig:intro}(c), our Anchor3DLane defines lane anchors as rays in the 3D space with given pitches and yaws.
Afterward, we first project them to corresponding 2D points in FV space using camera parameters, and then obtain their features by bilinear sampling. 
A simple classification head and a regression head are adopted to generate classification probabilities and 3D offsets from anchors respectively to make final predictions.
Unlike the information loss in IPM, sampling from original FV features retains richer context information around lanes, which helps estimate 3D information more accurately.
Moreover, our 3D lane anchors can be iteratively refined to sample more accurate features to better capture complex variations of 3D lanes.
Furthermore, Anchor3DLane can be easily extended to the multi-frame setting by projecting 3D anchors to adjacent frames with the assistance of camera poses between frames, which further improves performances over single-frame prediction.

In addition, we also utilize global constraints to refine the challenging distant parts due to low resolution.
The motivation is based on an intuitive insight that lanes in the same image appear to be parallel in most cases except for the fork lanes, i.e., distances between different point pairs on each lane pair are nearly consistent.
By applying a global equal-width optimization to non-fork lane pairs, we adjust 3D lane predictions to make the width of lane pairs consistent from close to far.
The lateral error of distant parts of lane lines can be further reduced through the above adjustment.

Our contributions are summarized as follows:
\begin{itemize}
\item We propose a novel Anchor3DLane framework that directly defines anchors in 3D space and regresses 3D lanes directly from FV without introducing BEV.
An extension to the multi-frame setting of Anchor3DLane is also proposed to leverage the well-aligned temporal information for further performance improvement.

\item We develop a global optimization method to utilize the equal-width properties of lanes for refinement.

\item Without bells and whistles, our Anchor3DLane outperforms previous BEV-based methods and achieves state-of-the-art performances on three popular 3D lane detection benchmarks.
\end{itemize}

\section{Related Works}
\subsection{2D Lane Detection}
2D lane detection~\cite{liu2021end, tabelini2021polylanenet, pan2018spatial, jin2022eigenlanes, yang2023lane} aims at obtaining the accurate shape and locations of 2D lanes in the images. 
Earlier works~\cite{aly2008real, he2004color, zhou2010novel, kim2008robust, wang2004lane} mainly focus on extracting low-level handcrafted features, such as edge and color information. However, these approaches often have complex feature extraction and post-processing designs and are less robust under changing scenarios.
With the development of deep learning, CNN-based methods have been explored recently and achieve notable performance. 
Segmentation-based methods~\cite{pan2018spatial, qin2020ultra, hou2019learning, neven2018towards} formulate 2D lane detection task as a per-pixel classification problem and typically focus on how to explore more effective and semantically informative features.
To make predictions more sparse and flexible, keypoint-based methods~\cite{qu2021focus, wang2022keypoint, ko2021key, rclane} model lane lines as sets of ordered keypoints and associate keypoints belonged to the same lane together by postprocessing.
Apart from the above methods, anchor-based methods~\cite{linecnn, laneatt, condlanenet, clrnet} are also popular in 2D lane detection task due to their conciseness and effectiveness. 
LineCNN~\cite{linecnn} first defines straight rays emitted from the image boundary to fit the shape of 2D lane lines and applies Non-Maximum Suppression (NMS) to keep only lanes with higher confidence.
LaneATT~\cite{laneatt} develops anchor-based feature pooling to extract features for the 2D anchors.
CLRNet~\cite{clrnet} learns to refine the initial anchors iteratively through the feature pyramid.

\subsection{3D Lane Detection}
Since projecting 2D lanes back into 3D space suffers from inaccuracy as well as less robustness, 3D lane detection task is proposed to predict lanes in 3D space end to end.
Some works utilize multiple sensors, such as stereo cameras~\cite{benmansour2008stereovision} and Lidar-camera~\cite{bai2018deep} to restore 3D information.
However, the collection and annotation cost of multi-sensor data is expensive, restricting the practical application of these methods. 
Therefore, monocular camera image based 3D lane detection~\cite{3dlanenet, genlanenet, clgo, once, 3dlanenet+} attracts more attention.

Due to the good geometric properties of lanes in the perspective of BEV, 3DLaneNet~\cite{3dlanenet} utilizes IPM to transform features from FV into BEV and then regresses the anchor offsets of lanes in the BEV space.
CLGo~\cite{clgo} transforms raw images into BEV images with the estimated camera pitches and heights and fits the lane lines by predicting polynomial parameters.
Since IPM relies heavily on the flat ground assumption, lanes represented in BEV space may be misaligned with 3D space in rough ground cases.
To this end, Gen-LaneNet~\cite{genlanenet} makes a distinction between the virtual top view generated by IPM and the true top view in 3D space for better space alignment.
Persformer~\cite{persformer} utilizes deformable attention to generate BEV features more adaptively and robustly.
SALAD~\cite{once} tries to get rid of BEV by decomposing 3D lane detection into 2D lane segmentation and dense depth estimation tasks. 
Different from the above methods, our Anchor3DLane defines anchors in the 3D space to explicitly model 3D lanes and bridge the gap between FV space and 3D space. 
The projection and sampling operations ensure the accuracy of anchor feature extraction, enabling effectively predicting 3D lanes directly from FV representations without introducing BEV.

\section{Method}
\label{sec:method}

\begin{figure*}[!t]
    \centering
    \includegraphics[width=\linewidth]{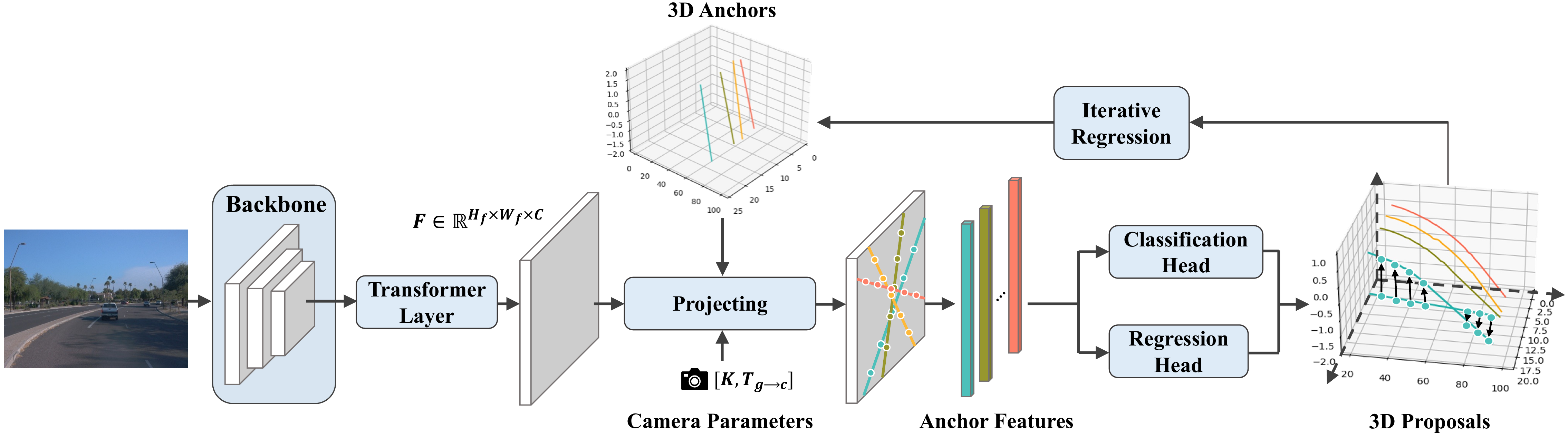}
    \caption{The overall architecture of Anchor3DLane. Given a front-viewed input image, a CNN backbone and a Transformer layer are adopted to first extract visual feature $\mathbf{F}$. 3D anchors are then projected to sample their features from $\mathbf{F}$ given camera parameters. Afterward, a classification head and a regression head are applied to make the final predictions. The lane predictions can also serve as new 3D anchors for iterative regression.}
    \label{fig:pipeline}
\end{figure*}

The overall architecture of our Anchor3DLane is illustrated in Figure~\ref{fig:pipeline}.
Given a front-viewed image $\mathbf{I} \in \mathbb{R}^{H\times W\times 3}$ as input, where $H$ and $W$ denote the height and width of the input image, a CNN backbone (e.g., ResNet-18~\cite{resnet}) is adopted to extract 2D visual features represented in FV space.
To enlarge the receptive field of the network, we further insert a single Transformer layer \cite{transformer} after the backbone to obtain the enhanced 2D feature map $\mathbf{F} \in \mathbb{R}^{H_f \times W_f \times C}$, where $H_f$, $W_f$, and $C$ represent the height, width and channel number of feature map respectively. 
3D anchors are then projected to this feature map $\mathbf{F}$ with the assistance of camera parameters, and the corresponding anchor features are sampled using bilinear interpolation.
Afterward, we apply a classification head and a regression head to the sampled anchor features to make predictions, with each head composed of several lightweight fully connected layers. Furthermore, the predictions can be regarded as new 3D anchors for iterative regression.
\begin{figure}[!t]
    \centering
    \includegraphics[width=\linewidth]{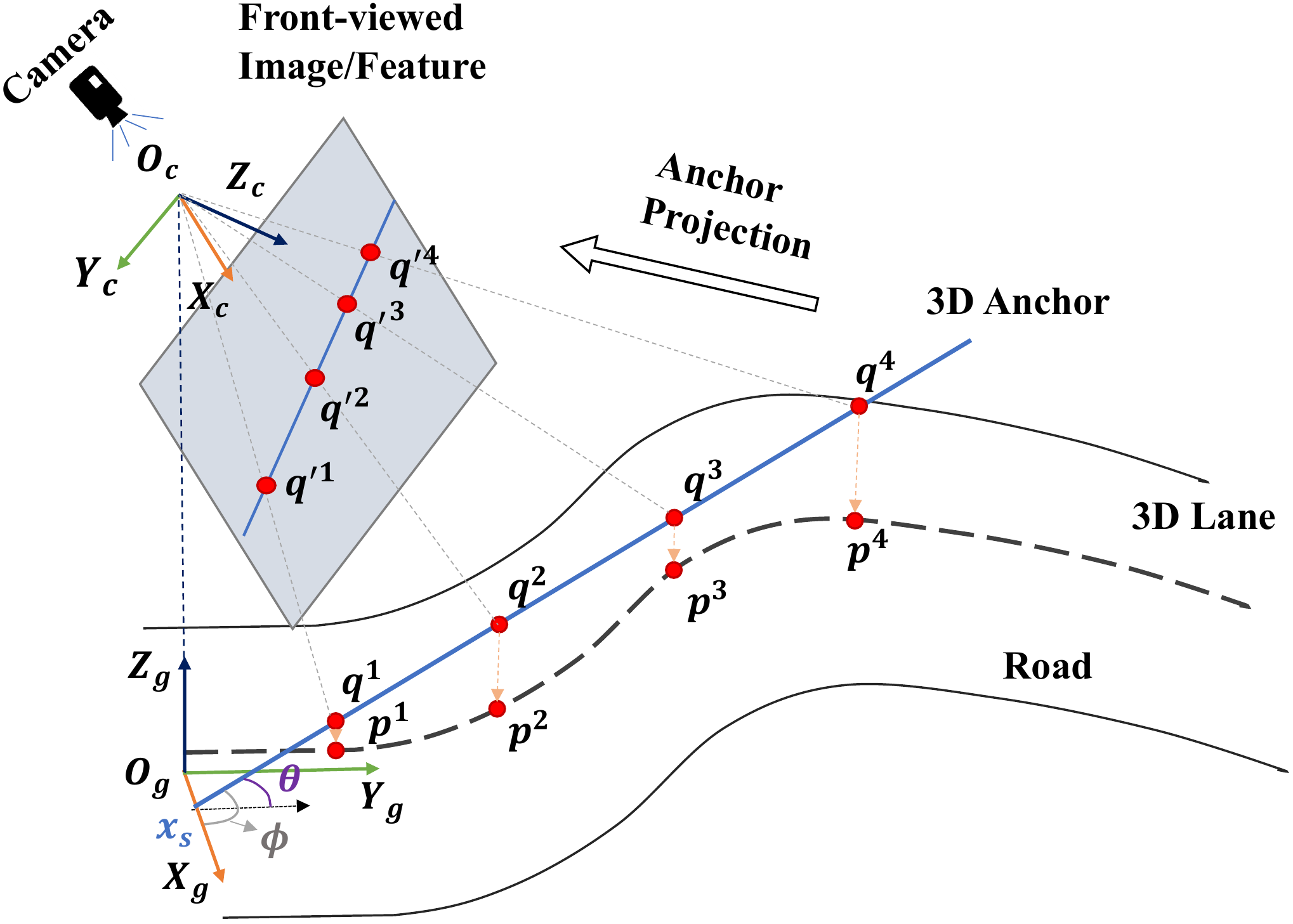}
    \caption{Illustration of 3D anchor and 3D lane in the ground coordinate system.}
    \label{fig:anchor}
    \vspace{-0.5cm}
\end{figure}

\subsection{3D Lane Representation}

We first revisit the representation of 3D lanes in this section.
As shown in Figure~\ref{fig:anchor}, two different coordinate systems are involved in our paper, including the camera coordinate system and the ground coordinate system.
The camera coordinate directly corresponds with the FV image and is a right-handed coordinate system defined by origin $O_c$ and $X_c, Y_c, Z_c$ axes, with $O_c$ located at the center of the camera and $Z_c$ pointing forward vertical to the camera plane.
3D lanes are commonly annotated in the ground coordinate system, of which the origin $O_g$ is set right below $O_c$, x-axis $X_g$ points positive to the right, y-axis $Y_g$ points positive forwards and z-axis $Z_g$ points positive upwards.
A 3D lane is described by 3D points with $N$ uniformly sampled y-coordinates $\mathbf{y}=\{y^k\}_{k=1}^N$.
Thus, we denote the $i$-th 3D lane as $\mathbf{G}_i=\{\mathbf{p}_i^k\}_{k=1}^N$ and its $k$-th point is represented as $\mathbf{p}_i^k=(x_i^k, y^k, z_i^k, vis_i^k)$, where the first $3$ elements denote the location of $\mathbf{p}_i^k$ in the ground coordinate system and the last one denotes the visibility of $\mathbf{p}_i^k$.
It is worth noting that we elaborate our method based on the ground coordinate system following the common practices adopted in previous works~\cite{3dlanenet, genlanenet}.
However, our Anchor3DLane is able to work in an arbitrary 3D coordinate system as long as camera calibration is available.

\vspace{0.3cm}
\subsection{Anchor3DLane}
\label{subsec:anchor}

\subsubsection{Representation of 3D Lane Anchors}
Our 3D lane anchors are defined in the same coordinate system as 3D lanes, i.e., ground coordinate, for ease of position regression.
As illustrated in Figure~\ref{fig:anchor}, a 3D anchor is a ray starting from $(x_s, 0, 0)$ with pitch $\theta$ and yaw $\phi$.
Similar to 3D lanes, we also sample $N$ points for each anchor by the same y-coordinates and represent the $j$-th 3D anchor by $\mathbf{A}_j=\{\mathbf{q}_j^k\}_{k=1}^N$, and its $k$-th point is denoted by $\mathbf{q}_j^k=(x_j^k, y^k, z_j^k)$.
Different from previous works~\cite{3dlanenet, persformer} that define anchors in the BEV plane, our 3D anchors have pitches to the ground and could fit the lane shape better.

\vspace{0.4cm}
\subsubsection{Anchor Projection and Feature Sampling}
To obtain features of 3D anchors, we first project them into the plane of FV feature $\mathbf{F}$ using camera parameters as shown in Figure~\ref{fig:anchor}.
Given an anchor $\mathbf{A}_j$, we take its $k$-th point $\mathbf{q}_j^k$ as an example to explain the projection operation and omit the subscript $j$ for simplicity as follows:
\vspace{0.3cm}
\begin{gather}
\label{eq:proj}
    \begin{bmatrix}
    \tilde{u}^k \\
    \tilde{v}^k \\
    d^k \\
    \end{bmatrix} = \mathbf{K} \mathbf{T}_{g \rightarrow c} \begin{bmatrix}
    x^k \\
    y^k \\
    z^k \\
    1
    \end{bmatrix}, \\
    u^k = W_f / W \cdot \frac{\tilde{u}^k}{d^k}, \\
    v^k = H_f / H \cdot \frac{\tilde{v}^k}{d^k},
\end{gather}
where $\mathbf{K} \in \mathbb{R}^{3\times 3}$ denotes camera intrinsic parameters, $\mathbf{T}_{g \rightarrow c} \in \mathbb{R}^{3\times 4}$ denotes the transform matrix from ground coordinate to camera coordinate, and $d^k$ denotes the depth of $\mathbf{q}^k$ to the camera plane.
Through the above formulations, $\mathbf{q}^k$ is projected to position $(u^k, v^k)$ in FV feature $\mathbf{F}$.
Finally, the feature of anchor $\mathbf{A}_j$ is obtained through bilinear interpolation within the neighborhood of the projected points and is represented as $\{\mathbf{F}_{(u^k, v^k)}\}_{k=1}^N$.

\vspace{0.3cm}
\subsubsection{3D Lane Prediction}
We concatenate features of points belonging to the same anchor as its feature representation.
Then we apply a classification head and a regression head to the anchor features for predicting classification probabilities $\mathbf{c}_j \in \mathbb{R}^L$, anchor points offsets $(\Delta \mathbf{x}_j \in \mathbb{R}^N, \Delta \mathbf{z}_j \in \mathbb{R}^N) = \{(\Delta x_j^k, \Delta z_j^k)\}_{k=1}^N$ and visibility of each point $\mathbf{vis}_j \in \mathbb{R}^N$ respectively, with $j \in [1, M]$.
$L$ and $M$ denote the numbers of lane types and 3D anchors respectively.
In this way, 3D lane proposals are generated as $\{\mathbf{P}_j=(\mathbf{c}_j, \mathbf{x}_j+\Delta \mathbf{x}_j, \mathbf{y}, \mathbf{z}_j+\Delta \mathbf{z}_j, \mathbf{vis}_j)\}_{j=1}^M$.
Furthermore, these 3D lane proposals can also serve as new anchors in the following iterative regression steps as illustrated in Figure~\ref{fig:pipeline}.
Through iterative regression, proposals can be refined progressively to better fit the lane shape.

During training, we associate $n$ nearest anchors to each ground-truth lane and the rest are defined as negative samples. Distance metric between ground-truth $\mathbf{G}_i$ and anchor $\mathbf{A}_j$ is calculated as follows:
\begin{equation}
\label{eq:dis}
    D(\mathbf{G}_i, \mathbf{A}_j) = \frac{\sum_{k=1}^N{vis_i^k \cdot \sqrt{(x_i^k - x_j^k)^2 + (z_i^k - z_j^k)^2}}}{ \sum_{k=1}^N{vis_i^k}}.
\end{equation}
This metric is also used in Non-Maximum Suppression (NMS) during inference to keep a reasonable number of proposals except that distances are calculated between visible parts of two proposals.

We adopt focal loss~\cite{focalloss} for training classification to balance the positive and negative proposals as follows:
\begin{equation}
    \mathcal{L}_{cls} = -\sum_{j=1}^M{\sum_{l=1}^L{\alpha^l (1-c_j^l)^\gamma \log{c_j^l}}},
\end{equation}
where $\alpha^l$ and $\gamma$ are the hyperparamters for focal loss.
The regression loss is only calculated between the positive proposals and their assigned ground-truth lanes following~\cite{genlanenet}:
\begin{equation}
\begin{aligned}
    \mathcal{L}_{reg} &= \sum_{i=1}^{M_p}{\sum_{k=1}^N{(\Vert \hat{vis}_i^k \cdot (x_i^k + \Delta x_i^k -\hat{x}_i^k) \Vert_1 }} \\
    &+ \sum_{i=1}^{M_p}{\sum_{k=1}^N{\Vert \hat{vis}_i^k \cdot (z_i^k + \Delta z_i^k -\hat{z}_i^k) \Vert_1)}} \\
    &+ \sum_{i=1}^{M_p}{\sum_{k=1}^N{\Vert \hat{vis}_i^k-vis_i^k \Vert_1}}.
\end{aligned}
\end{equation}
$M_p$ represents the total number of positive proposals.
Here we use $\hat{x}_i^k$, $\hat{z}_i^k$ and $\hat{vis}_i^k$ to denote the $x$, $z$ coordinates and visibility of the ground-truth lane points.

The total loss function of our Anchor3DLane is a combination of the above two losses with corresponding coefficients:
\begin{equation}
    \mathcal{L} = \lambda_{cls} \mathcal{L}_{cls} + \lambda_{reg} \mathcal{L}_{reg}.
\end{equation}

\subsection{Temporal Context Modeling}
\label{sec:temp}
Thanks to the design of 3D anchors, our Anchor3DLane can be easily extended to multi-frame 3D lane detection.
Given a 3D point $(x_t, y_t, z_t)$ in the $t$-th frame's ground coordinate system, we transform it to the $t'$-th frame's ground coordinate system with the following formulation:
\begin{equation}
\label{eq:temp_proj}
    \begin{bmatrix}
    x_{t'} \\
    y_{t'} \\
    z_{t'} \\
    \end{bmatrix} = \mathbf{T}_{g(t)\rightarrow g(t')} \begin{bmatrix}
    x_t \\
    y_t \\
    z_t \\
    1 \\
    \end{bmatrix},
\end{equation}
where $\mathbf{T}_{g(t)\rightarrow g(t')} \in \mathbb{R}^{3\times 4}$ denotes the transformation matrix from $t$-th frame to $t'$-th frame.
Together with Equation~\ref{eq:proj}, anchors defined in the current frame can be projected to previous frames for sampling their features.
For each anchor, we take its points from the current frame as query and points from previous frames as key and value to conduct cross-frame attention for feature aggregation.
By integrating the well-aligned anchor features from multiple frames, temporal context is incorporated into our Anchor3DLane to enlarge its perception range.

\subsection{Optimization with Equal-Width Constraint}
In most cases, lanes in 3D space are nearly parallel with each other, which is helpful in generating robust 3D estimations from monocular 2D images.
In this work, we leverage this geometry property of 3D lanes and formulate it as an equal-width constraint to adjust the x-coordinates of lane predictions.
Given two lane predictions $\mathbf{P}_{j}=\{\mathbf{p}_j^k\}_{k=1}^N$ and $\mathbf{P}_{j'}=\{\mathbf{p}_{j'}^k\}_{k=1}^N$, width between $\mathbf{P}_{j}$ and $\mathbf{P}_{j'}$ at point pair $\mathbf{p}_{j}^k$ and $\mathbf{p}_{j'}^k$ is calculated as:
\begin{equation}
w_{j,j'}^k = \vert \cos{\theta_j^k}(x_j^k + \tilde{\Delta} x_{j}^k - x_{j'}^k - \tilde{\Delta} x_{j'}^k) \vert,
\end{equation}
where $\tilde{\Delta} x_{j}^k$ and $\tilde{\Delta} x_{j'}^k$ denote the adjustment to $x_{j}^k$ and $x_{j'}^k$ to be optimized respectively and $\theta_j^k$ denotes the normal direction of the adjusted lane at $\mathbf{p}_j^k$.
The objective function of equal-width constraint is as follows:
\begin{equation}
\begin{aligned}
    \min_{\{\tilde{\Delta} \mathbf{x}_j\}_{j\in [1, Q]}} &\frac{1}{Q(Q-1)}\sum_{j=1}^{Q}{\sum_{j'=1, j' \neq j}^{Q}{\mathcal{L}(\mathbf{w}_{j,j'})}} \\
    &+ \alpha\frac{1}{Q} \sum_{j=1}^{Q}{\Vert \tilde{\Delta} \mathbf{x}_j \Vert_2},
\end{aligned}
\end{equation}
where
\begin{equation}
    \mathcal{L}(\mathbf{w}_{j,j'})=\sum_{k=1}^N{\vert w_{j, j'}^k-\frac{1}{N}\sum_{k'=1}^N{w_{j, j'}^{k'}}\vert}.
\end{equation}
We use $Q$ to denote the number of lane predictions after NMS. $\mathcal{L}(\mathbf{w}_{j,j'})$ restricts the width between $\mathbf{P}_j$ and $\mathbf{P}_{j'}$ to be consistent and the second item serves as a regularization to avoid the adjusted results being too far from the original predictions.
We run this optimization as a post-processing step to refine the prediction results of the network.

\begin{table*}[t]
\begin{center}
\resizebox{1\linewidth}{!}{
\begin{tabular}{c|c|cccccc}
\toprule
\textbf{Scene} & \textbf{Method} & \textbf{AP(\%)$\uparrow$} & \textbf{F1(\%)$\uparrow$} & \textbf{x err/C(m) $\downarrow$} & \textbf{x err/F(m)} $\downarrow$ & \textbf{z err/C(m) $\downarrow$} & \textbf{z err/F(m) $\downarrow$} \\
\hline
\multirow{7}{*}{Balanced Scene} & 3DLaneNet~\cite{3dlanenet} & 89.3 & 86.4 & 0.068 & 0.477 & 0.015 & \textbf{0.202} \\
& Gen-LaneNet~\cite{genlanenet} & 90.1 & 88.1 & 0.061 & 0.496 & 0.012 & 0.214 \\
& CLGo~\cite{clgo} & 94.2 & 91.9 & 0.061 & 0.361 & 0.029 & 0.250 \\
& PersFormer~\cite{persformer} & - & 92.9 & 0.054 & 0.356 & 0.010 & 0.234 \\
& GP~\cite{gp} & 93.8 & 91.9 & 0.049 & 0.387 & \textbf{0.008} & 0.213 \\
& \cellcolor{mygray}Anchor3DLane (Ours) & \cellcolor{mygray}\textbf{97.2} & \cellcolor{mygray}\textbf{95.6} & \cellcolor{mygray}0.052 & \cellcolor{mygray}0.306 & \cellcolor{mygray}0.015 & \cellcolor{mygray}0.223 \\
& \cellcolor{mygray}Anchor3DLane$\dagger$(Ours) & \cellcolor{mygray}97.1 & \cellcolor{mygray}95.4 & \cellcolor{mygray}\textbf{0.045} & \cellcolor{mygray}\textbf{0.300} & \cellcolor{mygray}0.016 & \cellcolor{mygray}0.223 \\
\hline
\multirow{7}{*}{Rare Subset} & 3DLaneNet~\cite{3dlanenet} & 74.6 & 72.0 & 0.166 & 0.855 & 0.039 & \textbf{0.521} \\
& Gen-LaneNet~\cite{genlanenet} & 79.0 & 78.0 & 0.139 & 0.903 & 0.030 & 0.539 \\
& CLGo~\cite{clgo} & 88.3 & 86.1 & 0.147 & 0.735 & 0.071 & 0.609 \\
& PersFormer~\cite{persformer} & - & 87.5 & 0.107 & 0.782 & 0.024 & 0.602 \\
& GP~\cite{gp} & 85.2 & 83.7 & 0.126 & 0.903 & \textbf{0.023} & 0.625 \\
& \cellcolor{mygray}Anchor3DLane (Ours) & \cellcolor{mygray}\textbf{96.9} & \cellcolor{mygray}\textbf{94.4} & \cellcolor{mygray}0.094 & \cellcolor{mygray}\textbf{0.693} & \cellcolor{mygray}0.027 & \cellcolor{mygray}0.579 \\
& \cellcolor{mygray}Anchor3DLane$\dagger$ (Ours) & \cellcolor{mygray}95.9 & \cellcolor{mygray}\textbf{94.4} & \cellcolor{mygray}\textbf{0.082} & \cellcolor{mygray}0.699 & \cellcolor{mygray}0.030 & \cellcolor{mygray}0.580 \\
\hline
\multirow{7}{*}{Visual Variations} & 3D-LaneNet~\cite{3dlanenet} & 74.9 & 72.5 & 0.115 & 0.601 & 0.032 & 0.230 \\
& Gen-LaneNet~\cite{genlanenet} & 87.2 & 85.3 & 0.074 & 0.538 & 0.015 & 0.232 \\
& CLGo~\cite{clgo} & 89.2 & 87.3 & 0.084 & 0.464 & 0.045 & 0.312 \\
& PersFormer~\cite{persformer} & - & 89.6 & 0.074 & 0.430 & 0.015 & 0.266 \\
& GP~\cite{gp} & 92.1 & 89.9 & 0.060 & 0.446 & \textbf{0.011 }& 0.235 \\
& \cellcolor{mygray}Anchor3DLane (Ours) & \cellcolor{mygray}\textbf{93.6} & \cellcolor{mygray}91.4 & \cellcolor{mygray}0.068 & \cellcolor{mygray}0.367 & \cellcolor{mygray}0.020 & \cellcolor{mygray}0.232 \\
& \cellcolor{mygray}Anchor3DLane$\dagger$ (Ours) & \cellcolor{mygray}92.5 & \cellcolor{mygray}\textbf{91.8} & \cellcolor{mygray}\textbf{0.047} & \cellcolor{mygray}\textbf{0.327} & \cellcolor{mygray}0.019 & \cellcolor{mygray}\textbf{0.219} \\
\bottomrule     
\end{tabular}}
\caption{Comparison with state-of-the-art methods on ApolloSim dataset with three different split settings. ``C'' and ``F'' are short for close and far respectively. $\dagger$ denotes iterative regression.}
\vspace{-0.5cm}
\label{tab:sota-apollo}
\end{center}
\end{table*}

\section{Experiments}

\subsection{Experimental Setting}
\subsubsection{Datasets and Evaluation Metrics}
We conduct experiments on three popular 3D lane detection benchmarks, including ApolloSim~\cite{genlanenet}, OpenLane~\cite{persformer}, and ONCE-3DLanes~\cite{once}.

\textbf{ApolloSim} is a photo-realistic synthetic dataset created with Unity 3D engine which contains 10.5K images from various virtual scenes, including highway, urban, residential, downtown, etc. In addition, the data is also diverse in daytime, weather conditions, traffic/obstacles, and road surface qualities.

\textbf{OpenLane} is a large-scale real-world 3D lane detection dataset constructed upon the Waymo Open dataset~\cite{waymo}.
It contains 200K frames and over 880K lanes are annotated. 
Camera intrinsics and extrinsics are provided for each frame.
All lanes are annotated including lanes in the opposite direction if no curbside in the middle. 
Categories and scene tags (e.g., weather and locations) are also provided.

\textbf{ONCE-3DLanes} is a real-world 3D lane detection dataset with $1$ million scenes.
It consists of $211$K images with labeled 3D lane points.
It covers different time periods (sunny, cloudy, rainy) and various regions (downtown, suburbs, highway, bridges, and tunnels).
Only camera intrinsics are provided in ONCE-3DLanes.

During the evaluation, the predictions and ground truth lanes are matched via minimum-cost flow where the pairwise cost is defined as the square root of the sum of pointwise Euclidean distance.
A prediction is considered as true positive if over $75\%$ of its points' distances to ground-truth points are less than a threshold, i.e., $1.5$m.
With the definition above, Average Precision (AP) and the maximum F1 score are further calculated, and x/z errors are counted separately at close ($0$-$40$m) and far ($40$-$100$m) ranges.
We report the results of F1 score, AP, and x/z-errors on ApolloSim dataset.
On OpenLane dataset, except for F1 score and x/z errors, we further report category accuracy which calculates the proportion of predictions whose categories are correctly predicted to all true positive predictions.
ONCE-3DLanes adopts a different way to match predictions and ground truth lanes.
The matching degree is firstly decided by IoU on the top-view plane and pairs above the threshold are further calculated with their unilateral Chamfer Distance ($CD$) as the matching error.
A true positive is counted when $CD$ is under the threshold. 
We report F1 score, precision, recall, and CD error for results on ONCE-3DLanes.

\begin{table*}[t]
\begin{center}
\resizebox{\linewidth}{!}{
\begin{tabular}{c|cccccc}
\toprule
\textbf{Method} & \textbf{F1(\%)$\uparrow$} & \textbf{Cate Acc(\%)$\uparrow$} & \textbf{x err/C(m) $\downarrow$} & \textbf{x err/F(m) $\downarrow$} & \textbf{z err/C(m) $\downarrow$} & \textbf{z err/F(m) $\downarrow$} \\ 
\hline
3D-LaneNet~\cite{3dlanenet} & 44.1 & - & 0.479 & 0.572 & 0.367 & 0.443 \\
GenLaneNet~\cite{genlanenet} & 32.3 & - & 0.591 & 0.684 & 0.411 & 0.521\\
PersFormer~\cite{persformer} & 50.5 & \textbf{92.3} & 0.485 & 0.553 & 0.364 & 0.431\\
\rowcolor{mygray}
Anchor3DLane (Ours) & 53.1 & 90.0 & 0.300 & 0.311 & \textbf{0.103} & 0.139 \\
\rowcolor{mygray}
Anchor3DLane$\dagger$ (Ours) & 53.7 & 90.9 & 0.276 & 0.311 & 0.107 & 0.138 \\
\rowcolor{mygray}
Anchor3DLane-T$\dagger$ (Ours) & \textbf{54.3} & 90.7 & \textbf{0.275} & \textbf{0.310} & 0.105 & \textbf{0.135} \\
\bottomrule
\end{tabular}}
\caption{Comparison with state-of-the-art methods on OpenLane validation set. $\dagger$ denotes iterative regression. Anchor3DLane-T denotes incorporating multi-frame information. ``Cate Acc'' means category accuracy.}
\vspace{-0.2cm}
\label{tab:sota-openlane}
\end{center}
\end{table*}

\begin{table*}[!htbp]
\begin{center}
\resizebox{\linewidth}{!}{
\begin{tabular}{c|ccccccc}
\toprule
\textbf{Method} & \textbf{All} & \textbf{Up \& Down} & \textbf{Curve} & \textbf{Extreme Weather} & \textbf{Night} & \textbf{Intersection} & \textbf{Merge \& Split} \\ 
\hline
3D-LaneNet~\cite{3dlanenet} & 44.1 & 40.8 & 46.5 & 47.5 & 41.5 & 32.1 & 41.7 \\
GenLaneNet~\cite{genlanenet} & 32.3 & 25.4 & 33.5 & 28.1 & 18.7 & 21.4 & 31.0 \\
PersFormer~\cite{persformer} & 50.5 & 42.4 & 55.6 & 48.6 & 46.6 & 40.0 & 50.7 \\
\rowcolor{mygray}
Anchor3DLane (Ours) & 53.1 & 45.5 & 56.2 & 51.9 & 47.2 & 44.2 & 50.5 \\
\rowcolor{mygray}
Anchor3DLane$\dagger$ (Ours) & 53.7 & 46.7 & 57.2 & 52.5 & 47.8 & 45.4 & 51.2 \\
\rowcolor{mygray}
Anchor3DLane-T$\dagger$ (Ours) & \textbf{54.3} & \textbf{47.2} & \textbf{58.0} & \textbf{52.7} & \textbf{48.7} & \textbf{45.8} & \textbf{51.7} \\
\bottomrule
\end{tabular}}
\caption{Comparison with state-of-the-art methods on OpenLane validation set. F1 score is presented for each scenario. $\dagger$ denotes iterative regression. Anchor3DLane-T denotes incorporating multi-frame information.}
\vspace{-0.5cm}
\label{tab:scene_openlane}
\end{center}
\end{table*}

\subsubsection{Implementation Details}
We choose ResNet-18~\cite{resnet} as the backbone of our Anchor3DLane.
To maintain feature resolution, we set the downsampling stride of its last two stages to $1$ and replace the $3\times 3$ convolutions with dilated convolutions.
The starting positions $x_s$ of 3D anchors are evenly placed along the x-axis with an interval of $1.3$m.
For each $x_s$, different yaws $\phi \in \{0^{\circ}, \pm 1^{\circ}, \pm 3^{\circ}, \pm 5^{\circ}, \pm 7^{\circ}, \pm 10^{\circ}, \pm 15^{\circ}, \pm 20^{\circ}, \pm 30^{\circ}\}$ and pitches $\theta \in \{ 0^{\circ}, \pm 1^{\circ}, \pm 2^{\circ}, \pm 5^{\circ}\}$ are set respectively.
The number of points $N$ for each anchor is set to $10$ for experiments on ApolloSim and ONCE and $20$ for OpenLane.
We resize the image to $360 \times 480$ before feeding it to the backbone and the shape of $\mathbf{F}$ is $45 \times 60 \times 64$.
During training, $\lambda_{cls}$ and $\lambda_{reg}$ are both set to $1$ and the number of positive proposals is set as $3$.
The distance threshold for NMS is $2$ during inference.
For multi-frame Anchor3DLane, each time we randomly select $1$ frame from the previous $5$ frames to interact with current frame during training, and select the first frame of the previous $5$ frames during inference.
Since car poses are only available in OpenLane dataset, we only conduct temporal experiments on this dataset.
We use Adam optimizer~\cite{kingma2014adam} with weight decay set as $1e^{-4}$, and set the initial learning rate to $1e^{-4}$.
Step learning rate decay is used during training. 
$\alpha^l$ is set to $0.5$ and  $\gamma$ is set to $2$ for focal loss.
More details about our Anchor3DLane are included in supplementary materials.

\subsection{Quantitative Results}
\subsubsection{Results on ApolloSim}
Table~\ref{tab:sota-apollo} shows the experimental results under three different split settings of the ApolloSim dataset, including balanced scene, rare subset and visual variations.
We report the results of both our original Anchor3DLane and Anchor3DLane with iterative regression optimized with equal-width constraint.
It is shown that our original Anchor3DLane outperforms previous methods with large margins on AP and F1 score on all the three splits with simple design, i.e., $+3.0\%$ AP and $+2.7\%$ F1 score on balanced scene, $+8.6\%$ AP and $+6.9\%$ F1 score on rare subset, $+2.4\%$ F1 score and $+1.5\%$ AP on visual variations, showing the superiority of our method. 
Our Anchor3DLane also achieves comparable or lower x/z errors compared with previous methods, especially for x error far, indicating regressing over 3D anchors have greater advantages for distant predictions.
Furthermore, by iteratively regressing over the proposals predicted by Anchor3DLane, x/z errors can be further reduced to better fit the shape of 3D lanes.
\vspace{-0.3cm}

\subsubsection{Results on OpenLane}
We present the experimental results of our method optimized with the equal-width constraint on OpenLane dataset in Table~\ref{tab:sota-openlane}.
Our original Anchor3DLane outperforms PersFormer by $2.6\%$ F1 score improvement.
Moreover, our method achieves much more precise predictions than PersFormer, i.e., $-0.185$m on x error close, $-0.242$m on x error far, $-0.261$m on z error close, and $-0.292$m on z error far respectively, which are crucial for driving safety.
The gap in x/z errors indicates that under real scenarios with diverse conditions, directly sampling features from FV representation could maintain more environment context information, thus producing more precise predictions.
By incorporating iterative regression and temporal information in Anchor3DLane, the overall performances can be further boosted.
In Table~\ref{tab:scene_openlane}, we compare with previous methods under different scenarios and report F1 score for each scenario.
Our method produces much better performance in Up\&Down scenarios, showing the advantage of 3D anchor regression in uneven ground.
It is also worth noting that we adopt a lightweight CNN, i.e., ResNet-18 as the backbone of Anchor3DLane, which still outperforms PersFormer with a larger backbone, i.e., EfficientNet-B7~\cite{efficientnet}.
\vspace{-0.5cm}

\subsubsection{Results on ONCE-3DLanes}
In Table~\ref{tab:once}, we present the experimental results on the ONCE-3DLanes dataset. 
Since camera extrinsics are not available in ONCE-3DLanes, we define the 3D anchors in the camera coordinate system and make predictions in the same space.
Our method also achieves state-of-the-art performances on this dataset.
Compared with PersFormer, our Anchor3DLane still produces a higher F1 score and reduces CD error by $18.9\%$ relatively, which indicates that 3D anchors are able to adapt different 3D coordinate systems.

\begin{table}[!htbp]
\begin{center}
\resizebox{\linewidth}{!}{
\begin{tabular}{c|cccc}
\toprule
\textbf{Method} & \textbf{F1(\%)$\uparrow$} & \textbf{P(\%)$\uparrow$ }& \textbf{R(\%)$\uparrow$} & \textbf{CD Error(m)$\downarrow$} \\ 
\hline
3D-LaneNet~\cite{3dlanenet} & 44.73 & 61.46 & 35.16 & 0.127 \\
Gen-LaneNet~\cite{genlanenet} & 45.59 & 63.95 & 35.42 & 0.121\\
SALAD~\cite{once} & 64.07 & 75.90 & 55.42 & 0.098 \\
PersFormer~\cite{persformer} & 74.33 & 80.30 & 69.18 & 0.074\\
\rowcolor{mygray}
Anchor3DLane (Ours) & 74.44 & 80.50 & 69.23 & 0.064 \\
\rowcolor{mygray}
Anchor3DLane$\dagger$ (Ours) & \textbf{74.87} & \textbf{80.85} & \textbf{69.71} & \textbf{0.060} \\
\bottomrule     
\end{tabular}}
\caption{Comparison with state-of-the-art methods on ONCE-3DLanes validation set. Results under $\tau_{CD}=0.3$ are displayed here. $\dagger$ denotes iterative regression. ``P'' and ``R'' are short for precision and recall respectively.}
\vspace{-0.6cm}
\label{tab:once}
\end{center}
\end{table}

\vspace{-0.3cm}
\subsubsection{Ablation Study}
In this section, we follow previous work~\cite{persformer} to conduct most ablation studies on OpenLane-300, which is a subset of OpenLane.
As for feature sampling experiments, we present the results on the original OpenLane to verify the effectiveness of our method.
More ablation studies and qualitative results are included in the supplementary materials.

\begin{table}[!htbp]
\begin{center}
\resizebox{\linewidth}{!}{
\begin{tabular}{c|c|ccccc}
\toprule
\textbf{Input} & \textbf{Feat} & \textbf{F1(\%)} & \textbf{x err/C(m)}& \textbf{x err/F(m)} & \textbf{z err/C(m)} & \textbf{z err/F(m)} \\
\hline 
BEV & BEV & 47.6 & 0.466 & 0.421 & 0.119 & 0.170 \\
FV  & BEV & 47.6 & 0.443 & 0.446 & 0.118 & 0.160 \\
FV & FV & \textbf{53.1} & \textbf{0.300} & \textbf{0.31} & \textbf{0.103} & \textbf{0.139} \\
\bottomrule     
\end{tabular}}
\caption{Comparison between sampling anchor features from BEV features and FV features.}
\vspace{-0.5cm}
\label{tab:bev}
\end{center}
\end{table}

\textbf{Sampling anchor features from FV features.} 
To illustrate the superiority of FV features, we compare the performances of extracting anchor features from FV features and BEV features. The results are shown in Table~\ref{tab:bev}.
We explore different ways of obtaining BEV features, including warping FV image to BEV image (line 1) and warping FV feature to BEV feature (line 2), and keep the other settings same as our original Anchor3DLane.
Results show that sampling anchor features from FV features produces the best F1 score and x/z errors, especially for x errors, where more than $10$cm gap exists between FV anchor features and BEV anchor features.
The above performance gap indicates that the context information contained in raw FV features is beneficial for accurate lane predictions.

\begin{table}[!htbp]
\begin{center}
\resizebox{\linewidth}{!}{
\begin{tabular}{c|ccccc}
\toprule
\textbf{Iter} & \textbf{F1(\%)} & \textbf{x err/C(m)}& \textbf{x err/F(m)} & \textbf{z err/C(m)} & \textbf{z err/F(m)} \\
\hline 
1 & 54.8 & 0.318 & 0.349 & \textbf{0.101} & \textbf{0.147} \\ 
2 & 56.3 & \textbf{0.287} & 0.335 & 0.103 & 0.152 \\
3 & \textbf{57.0} & \textbf{0.287} & \textbf{0.327} & 0.104 & 0.148 \\
\bottomrule     
\end{tabular}}
\caption{Ablation study on the steps of iterative regression.}
\label{tab:iter}
\vspace{-0.5cm}
\end{center}
\end{table}

\textbf{Steps of iterative regression.}
Table~\ref{tab:iter} presents the results of different steps of iterative regression for Anchor3DLane.
Compared with no iterative regression, $2$ iterations produces relatively large performance improvements.
More steps of iterative regression can further reduce lateral errors as well as elevate F1 score by refining the shape of proposals progressively.

\begin{table}[!htbp]
\begin{center}
\resizebox{\linewidth}{!}{
\begin{tabular}{c|ccccc}
\toprule
\textbf{Method} & \textbf{F1(\%)} & \textbf{x err/C(m)}& \textbf{x err/F(m)} & \textbf{z err/C(m)} & \textbf{z err/F(m)} \\
\hline
w/o Temporal & 54.8 & 0.318 & 0.349 & 0.101 & 0.147 \\
Linear Fusion & 54.9 & 0.322 & 0.343 & 0.102 & 0.148 \\
Weighted Sum & \textbf{55.8} & 0.320 & 0.346 & 0.101 & 0.150 \\
Attention & 55.2 & \textbf{0.308} & \textbf{0.330} & \textbf{0.099} & \textbf{0.145} \\
\bottomrule     
\end{tabular}}
\caption{Ablation study on temporal integration methods.}
\vspace{-0.6cm}
\label{tab:temp}
\end{center}
\end{table}

\textbf{Temporal integration methods}.
In this section, we explore different methods to integrate anchor features of multiple frames.
Besides the cross-frame attention that we mentioned in Section~\ref{sec:temp}, we also try \textit{linear fusion} which concatenates features of the same anchor along their channels and fuses them with a linear layer, and \textit{weighted sum} which learns to predict a group of weights for each y-coordinate to fuse features of the same anchor elementwisely,
As shown in Table~\ref{tab:temp}, comparing with the baseline, incorporating temporal information into Anchor3DLane can improve the overall performance significantly due to the richer context information obtained from previous frames.
Weighted sum produces better results than linear fusion, indicating that dynamic weights are necessary for different points at different distances.
Although weighted sum achieves a better F1 score compared with single frame setting, x/z errors increase at the same time.
Among the $3$ integration methods, cross-frame attention, which aggregates anchor features with more anchor points from previous frames, improves both F1 score and x errors and achieves the best performance balance.

\begin{table}[htbp]
\begin{center}
\resizebox{0.8\linewidth}{!}{
\begin{tabular}{c|ccc}
\toprule
\textbf{Method} & \textbf{F1(\%)} & \textbf{x err/C(m) } & \textbf{x err/F(m)} \\ 
\hline 
w/o EWC & 54.8 & \textbf{0.318} & 0.349 \\
w/ EWC & \textbf{55.0} & \textbf{0.318} & \textbf{0.337} \\
\bottomrule     
\end{tabular}}
\caption{Ablation study on Equal-Width Constraint (EWC).}
\vspace{-0.5cm}
\label{tab:ewc}
\end{center}
\end{table}

\textbf{Effect of equal-width constraint}.
We also illustrate the comparison between predictions without and with equal-width constraint optimization.
As shown in Table~\ref{tab:ewc}, by applying the equal-width constraint to the lane predictions, errors of the distant parts of the lane lines can be further reduced by restricting them to have the same width as the close parts.
More visualization results of this constraint can be found in the supplementary materials.

\section{Conclusion}
In this work, we propose a novel Anchor3DLane framework for 3D lane detection which bypasses the transformation to BEV space and predicts 3D lanes from FV directly.
By defining anchors in the 3D space and projecting them to the FV features, accurate anchor features are sampled for lane prediction.
We further extend our Anchor3DLane to the multi-frame setting to incorporate temporal information, which improves performances due to the enriched context.
In addition, a global equal-width optimization method is proposed to utilize the parallel property of lanes for refinement.
Experimental results show that our Anchor3DLane outperforms previous methods on three 3D lane detection benchmarks with a simple architecture.

\section{Acknowledgments}
This research was supported in part by National Key R\&D Program of China (2022ZD0115502), National Natural Science Foundation of China (No. 62122010), Zhejiang Provincial Natural Science Foundation of China under Grant No. LDT23F02022F02, Key Research and Development Program of Zhejiang Province under Grant No. 2022C01082.

{\small
\bibliographystyle{ieee_fullname}
\bibliography{ref}
}
\clearpage
\begin{appendices}

\section{Implementation Details}
\textbf{ApolloSim}.
We resample $10$ points for the ApolloSim dataset at y-coordinates of \{5, 10, 15, 20, 30, 40, 50, 65, 80, 100\}.
The training batch size is set to $16$.
We train Anchor3DLane on this dataset with one NVIDIA RTX 2080 Ti GPU for $50,000$ iterations and decay the learning rate at the $45,000$-th iteration by $10$ times.

\textbf{OpenLane}.
We resample $20$ points for the OpenLane dataset at y-coordinates of \{5, 10, 15, 20, 25, 30, 35, 40, 45, 50, 55, 60, 65, 70, 75, 80, 85, 90, 95, 100\}.
The training batch size is set to $64$.
We train Anchor3DLane on this dataset with eight NVIDIA RTX 2080 Ti GPUs for $60,000$ iterations and decay the learning rate at the $50,000$-th iteration by $10$ times.

\textbf{ONCE-3DLanes}.
We resample $10$ points for the ONCE-3DLanes dataset at y-coordinates of \{2, 5, 8, 10, 15, 20, 25, 30, 40, 50\}.
Since experiments are conducted in the camera coordinate system where the origin is above the ground, the starting positions of 3D anchors are set at $(x_s, 0, -1.5m)$.
Other training settings are the same as those on the OpenLane dataset as mentioned above.

\section{Quantitative Results}
\subsection{The Range of Training Frames}

\begin{table}[htbp]
\begin{center}
\resizebox{1.0\linewidth}{!}{
\begin{tabular}{c|ccccc}
\toprule
\textbf{Frame Range} & \textbf{F1(\%)} & \textbf{x err/C(m) } & \textbf{x err/F(m)} & \textbf{z err/C(m)} & \textbf{z err/F(m)} \\ 
\hline 
3 frames & 55.0 & \textbf{0.306} & \textbf{0.326} & \textbf{0.099} & 0.148 \\
5 frames & 55.2 & 0.308 &  0.330 & \textbf{0.099} & \textbf{0.145} \\
7 frames & \textbf{56.1} & 0.312 & 0.335 & 0.101 & 0.150 \\
\bottomrule     
\end{tabular}}
\caption{Ablation study on the range of training frames.}
\label{tab:temp}
\end{center}
\end{table}

For temporal context modeling, we sample one frame from different ranges of previous frames to aggregate its feature to the current frame during training.
The first frame of the previous 5 frames is sampled during inference.
As shown in Table~\ref{tab:temp}, the F1 score increases as the frame range becomes larger, indicating that aggregating information from farther frames yields a better estimation for the current frame.

\subsection{Computational Cost Analysis}
\begin{table}[htbp]
\begin{center}
\resizebox{1.0\linewidth}{!}{
\begin{tabular}{c|cccc}
\toprule
\textbf{Method} & \textbf{F1 Score(\%)} & \textbf{FLOPs} & \textbf{Param} & \textbf{FPS} \\
\hline 
PersFormer~\cite{persformer} & 50.5 &  572.4G & 54.9M & 5.58 \\
Anchor3DLane (ours) & 53.1 & 38.1G & 12.2M & 87.29 \\
Anchor3DLane$\dagger$ (ours) & 53.7 & 42.4G & 13.2M & 73.73 \\
Anchor3DLane-T$\dagger$ (ours) & 54.3 & 82.3G & 13.3M & 30.22 \\
\bottomrule     
\end{tabular}}
\caption{Comparison of computational cost and F1 score on OpenLane validation set. $\dagger$ denotes iterative regression. Anchor3DLane-T denotes incorporating multi-frame information.}
\label{tab:cost}
\end{center}
\end{table}

We report the computational cost comparison in Table~\ref{tab:cost}.
Our Anchor3DLane achieves a higher F1 score on the OpenLane dataset with much fewer FLOPs and parameters compared with PersFormer~\cite{persformer}.
The inference speeds (FPS) of these methods are measured using the code released by PersFormer on a single 2080 Ti GPU.
Our original Anchor3DLane achieves nearly $16$ times faster inference speed than PersFormer.
By adopting iterative regression and temporal context modeling, the F1 score is further improved, while the inference speed decreases but is still much faster than PersFormer.
These results demonstrate our Anchor3DLane is both effective and efficient.

\subsection{Experimental Results with EfficientNet}
To verify the adaptability and performance potential of our method, we further conduct experiments with EfficientNet-B3~\cite{efficientnet} to compare with PersFormer which adopts EfficientNet-B7 as the backbone.
Results are shown in Table~\ref{tab:eff-openlane}, Table~\ref{tab:scene_openlane_eff} and Table~\ref{tab:eff-once}.
On OpenLane dataset, utilizing EfficientNet-B3 as the backbone could boost the performance of our Anchor3DLane from $53.1\%$ F1 score to $56.0\%$ and reduce the x/z errors at the same time, indicating that our method adapts well to stronger backbones.

\begin{table*}[t]
\begin{center}
\resizebox{\linewidth}{!}{
\begin{tabular}{c|c|cccccc}
\toprule
\textbf{Method} & \textbf{Backbone} & \textbf{F1(\%)$\uparrow$} & \textbf{Cate Acc(\%)$\uparrow$} & \textbf{x err/C(m) $\downarrow$} & \textbf{x err/F(m) $\downarrow$} & \textbf{z err/C(m) $\downarrow$} & \textbf{z err/F(m) $\downarrow$} \\ 
\hline
3D-LaneNet~\cite{3dlanenet} & VGG-16~\cite{simonyan2014very} & 44.1 & - & 0.479 & 0.572 & 0.367 & 0.443 \\
GenLaneNet~\cite{genlanenet} & ERFNet~\cite{romera2017erfnet} & 32.3 & - & 0.591 & 0.684 & 0.411 & 0.521\\
PersFormer~\cite{persformer} & EfficientNet-B7~\cite{efficientnet} & 50.5 & \textbf{92.3} & 0.485 & 0.553 & 0.364 & 0.431\\
\rowcolor{mygray}
Anchor3DLane (Ours) & ResNet-18~\cite{resnet} & 53.1 & 90.0 & 0.300 & \textbf{0.311} & \textbf{0.103} & 0.139 \\
\rowcolor{mygray}
Anchor3DLane (Ours) & EfficientNet-B3 & \textbf{56.0} & 89.9 & \textbf{0.293} & 0.317 & \textbf{0.103} & \textbf{0.130} \\
\bottomrule
\end{tabular}}
\caption{Comparison with state-of-the-art methods on OpenLane validation set with stronger backbone.}
\label{tab:eff-openlane}
\end{center}
\end{table*}

\begin{table*}[!htbp]
\begin{center}
\resizebox{\linewidth}{!}{
\begin{tabular}{c|c|ccccccc}
\toprule
\textbf{Method} & \textbf{Backbone} & \textbf{All} & \textbf{Up \& Down} & \textbf{Curve} & \textbf{Extreme Weather} & \textbf{Night} & \textbf{Intersection} & \textbf{Merge \& Split} \\ 
\hline
3D-LaneNet~\cite{3dlanenet} & VGG-16 & 44.1 & 40.8 & 46.5 & 47.5 & 41.5 & 32.1 & 41.7 \\
GenLaneNet~\cite{genlanenet} & ERFNet &  32.3 & 25.4 & 33.5 & 28.1 & 18.7 & 21.4 & 31.0 \\
PersFormer~\cite{persformer} & EfficientNet-B7 & 50.5 & 42.4 & 55.6 & 48.6 & 46.6 & 40.0 & 50.7 \\
\rowcolor{mygray}
Anchor3DLane (Ours) & ResNet-18 & 53.1 & 45.5 & 56.2 & 51.9 & 47.2 & 44.2 & 50.5 \\
\rowcolor{mygray}
Anchor3DLane (Ours) & EfficientNet-B3 & \textbf{56.0} & \textbf{50.3} & \textbf{59.1} & \textbf{53.6}  & \textbf{52.8} & \textbf{47.4} & \textbf{53.3} \\ 
\bottomrule
\end{tabular}}
\caption{Comparison with state-of-the-art methods on OpenLane validation set with stronger backbone. F1 score is presented for each scenario. }
\label{tab:scene_openlane_eff}
\end{center}
\end{table*}

\begin{table*}[!htbp]
\begin{center}
\resizebox{0.8\linewidth}{!}{
\begin{tabular}{c|c|cccc}
\toprule
\textbf{Method} & \textbf{Backbone} & \textbf{F1 Score(\%)$\uparrow$} & \textbf{Precision(\%)$\uparrow$ }& \textbf{Recall(\%)$\uparrow$} & \textbf{CD Error(m)$\downarrow$} \\ 
\hline
3D-LaneNet~\cite{3dlanenet} & VGG-16 & 44.73 & 61.46 & 35.16 & 0.127 \\
Gen-LaneNet~\cite{genlanenet} & ERFNet & 45.59 & 63.95 & 35.42 & 0.121\\
SALAD~\cite{once} & SegFormer~\cite{segformer} & 64.07 & 75.90 & 55.42 & 0.098 \\
PersFormer~\cite{persformer} & EfficientNet-B7 & 74.33 & 80.30 & 69.18 & 0.074\\
\rowcolor{mygray}
Anchor3DLane (Ours) & ResNet-18 & 74.44 & 80.50 & 69.23 & \textbf{0.064} \\
\rowcolor{mygray}
Anchor3DLane (Ours) & EfficientNet-B3 &  \textbf{75.02} & \textbf{83.22} & 68.29 & \textbf{0.064} \\
\bottomrule     
\end{tabular}}
\caption{Comparison with state-of-the-art methods on ONCE-3DLanes validation set with stronger backbone.}
\label{tab:eff-once}
\end{center}
\end{table*}

\section{Qualitative Results}
\begin{figure*}[!htbp]
    \centering
    \includegraphics[width=0.85\linewidth]{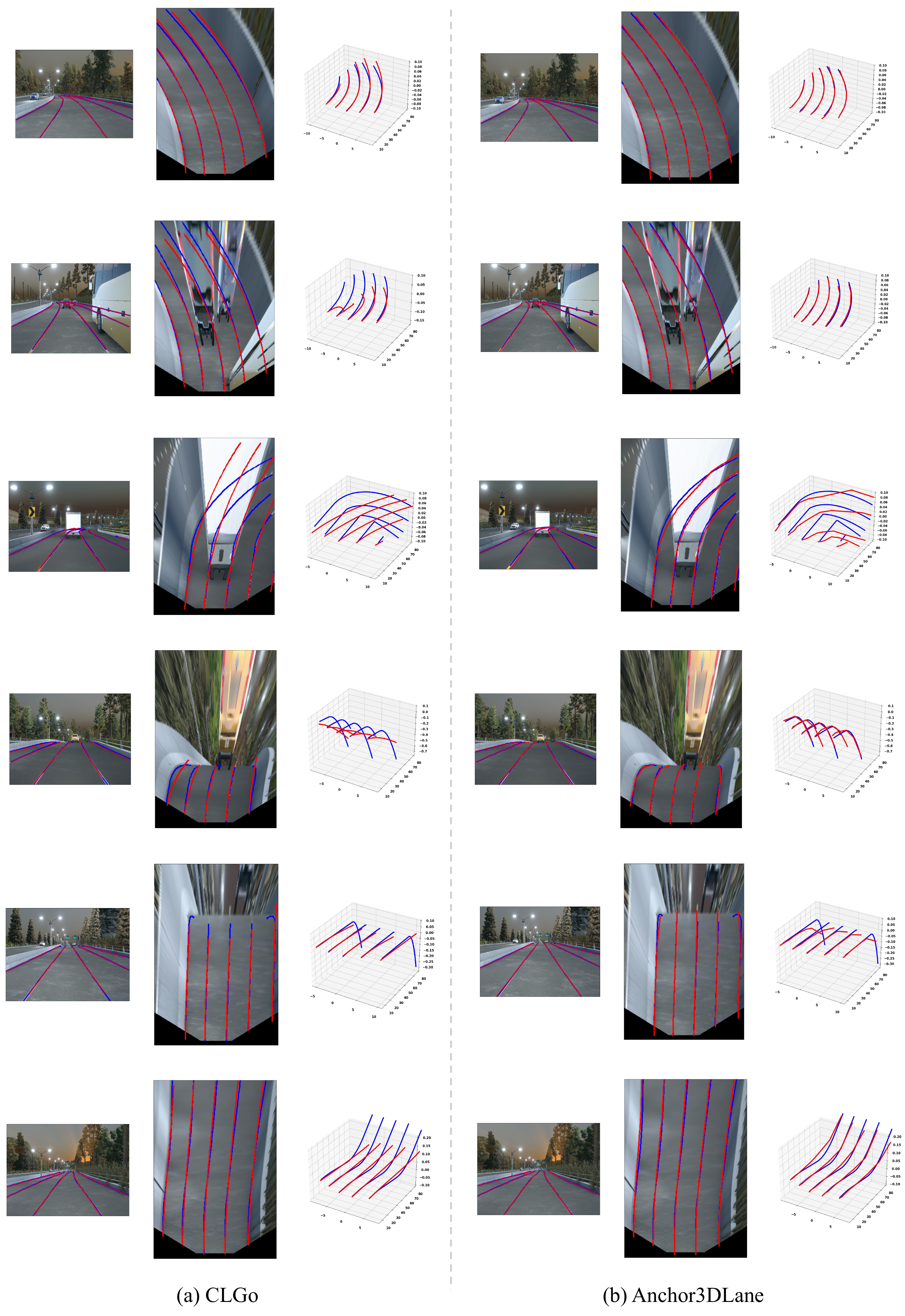}
    \caption{Comparison between CLGo~\cite{clgo} and our Anchor3DLane on the ApolloSim dataset. (a): Qualitative results of CLGo. (b): Qualitative results of our Anchor3DLane. \textcolor{blue}{Blue}:Ground-truth. \textcolor{red}{Red}: Prediction.}
    \label{fig:vis_apollo}
\end{figure*}

\textbf{ApolloSim.}
We compare our Anchor3DLane with CLGo~\cite{clgo} on the ApolloSim dataset and the results are included in Figure~\ref{fig:vis_apollo}.
Our method has better lateral predictions in the distant parts when lanes turn in the distance (row $2$ and row $3$).
In addition, when encountering uphill (row $6$) or downhill (row $4$ and row $5$), our method can better capture the height changes than CLGo, which demonstrates the superiority of directly regressing 3D anchors for 3D lanes.

\begin{figure*}[!htbp]
    \centering
    \includegraphics[width=0.9\linewidth]{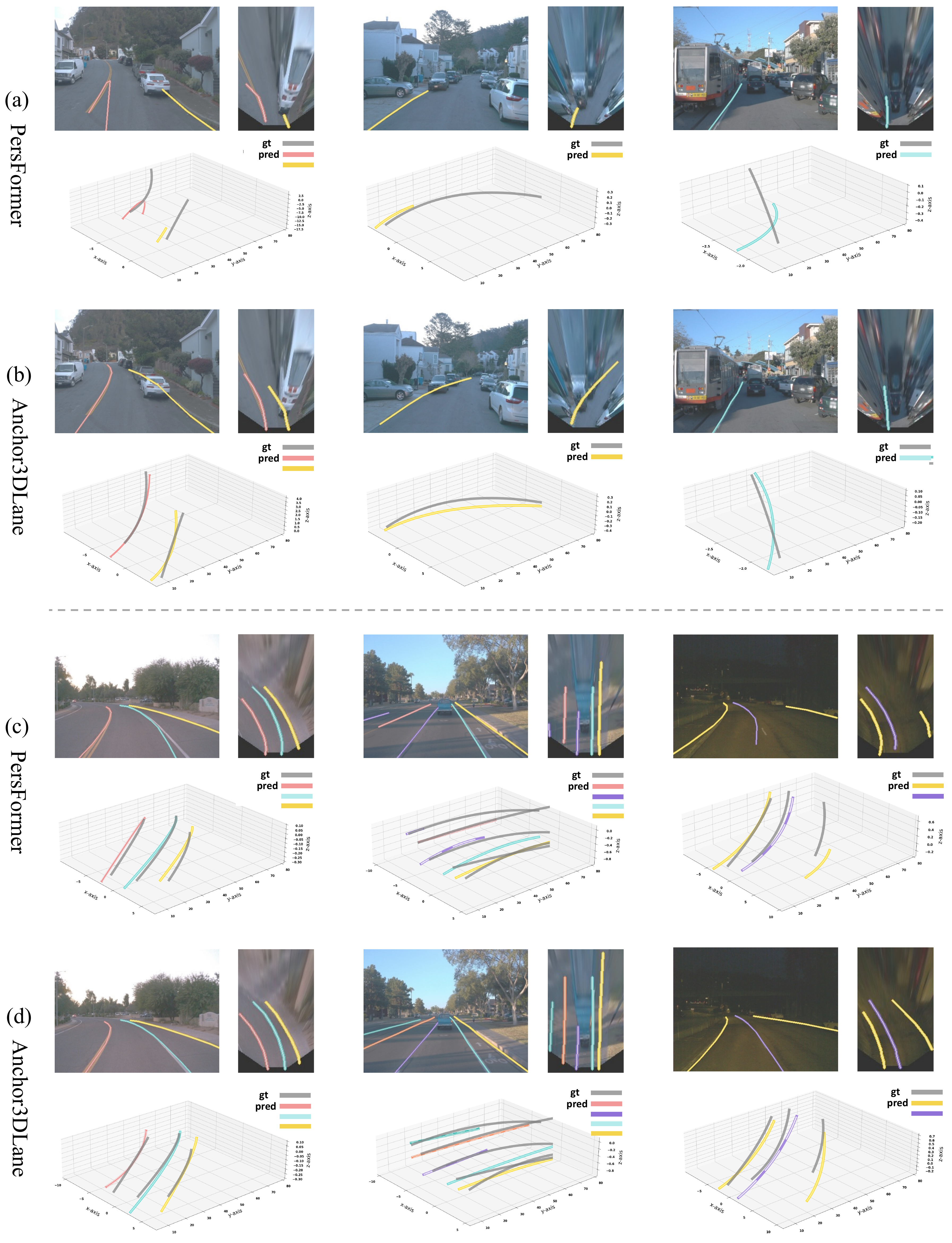}
    \caption{Comparison between PersFormer~\cite{persformer} and our Anchor3DLane on the OpenLane dataset. (a)(c): Qualitative results of PersFormer. (b)(d): Qualitative results of our Anchor3DLane.}
    \label{fig:vis_openlane}
\end{figure*}

\textbf{OpenLane.}
We also compare with PersFormer~\cite{persformer} on the OpenLane dataset in Figure~\ref{fig:vis_openlane}.
Our Anchor3DLane can better recover the whole lanes occluded by vehicles as shown in column $2$ of Figure~\ref{fig:vis_openlane} (a) and (b).

\begin{figure*}[!htbp]
    \centering
    \includegraphics[width=0.95\linewidth]{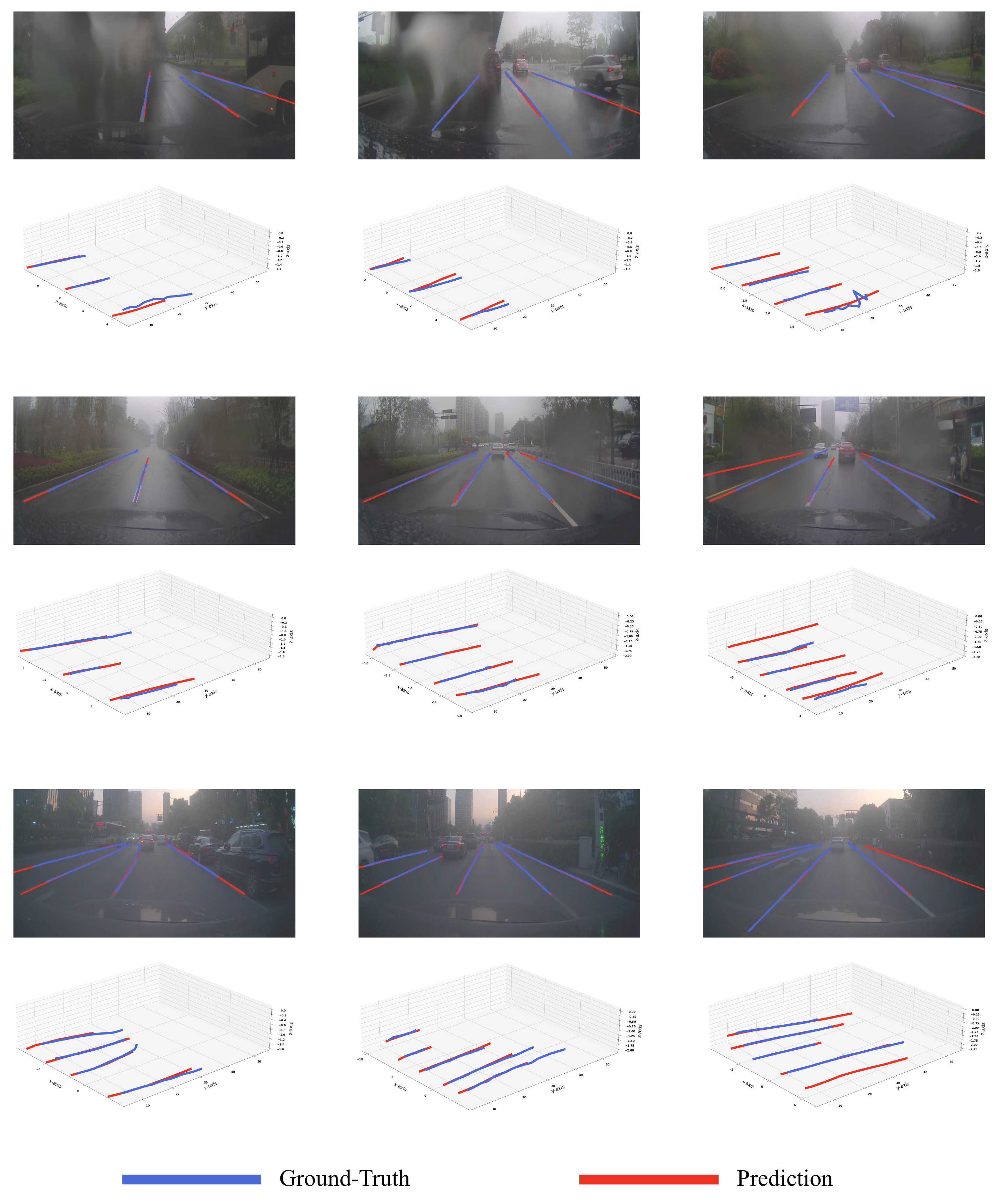}
    \caption{Qualitative results on the ONCE-3DLanes dataset.}
    \label{fig:vis_once}
\end{figure*}

\textbf{ONCE-3DLanes.}
In Figure~\ref{fig:vis_once}, we show the qualitative results of our Anchor3DLane on the ONCE-3DLanes dataset.
Our method performs well in different scenes, such as bad weather like rainy days (column $1$ of row $1$ and row $2$).
Since the 2D annotations of ONCE-3DLanes are generated by the lane detection model, annotations of some cases are inaccurate or incomplete but our method still produces fine predictions as shown in column $3$.

\begin{figure*}[!htbp]
    \centering
    \includegraphics[width=\linewidth]{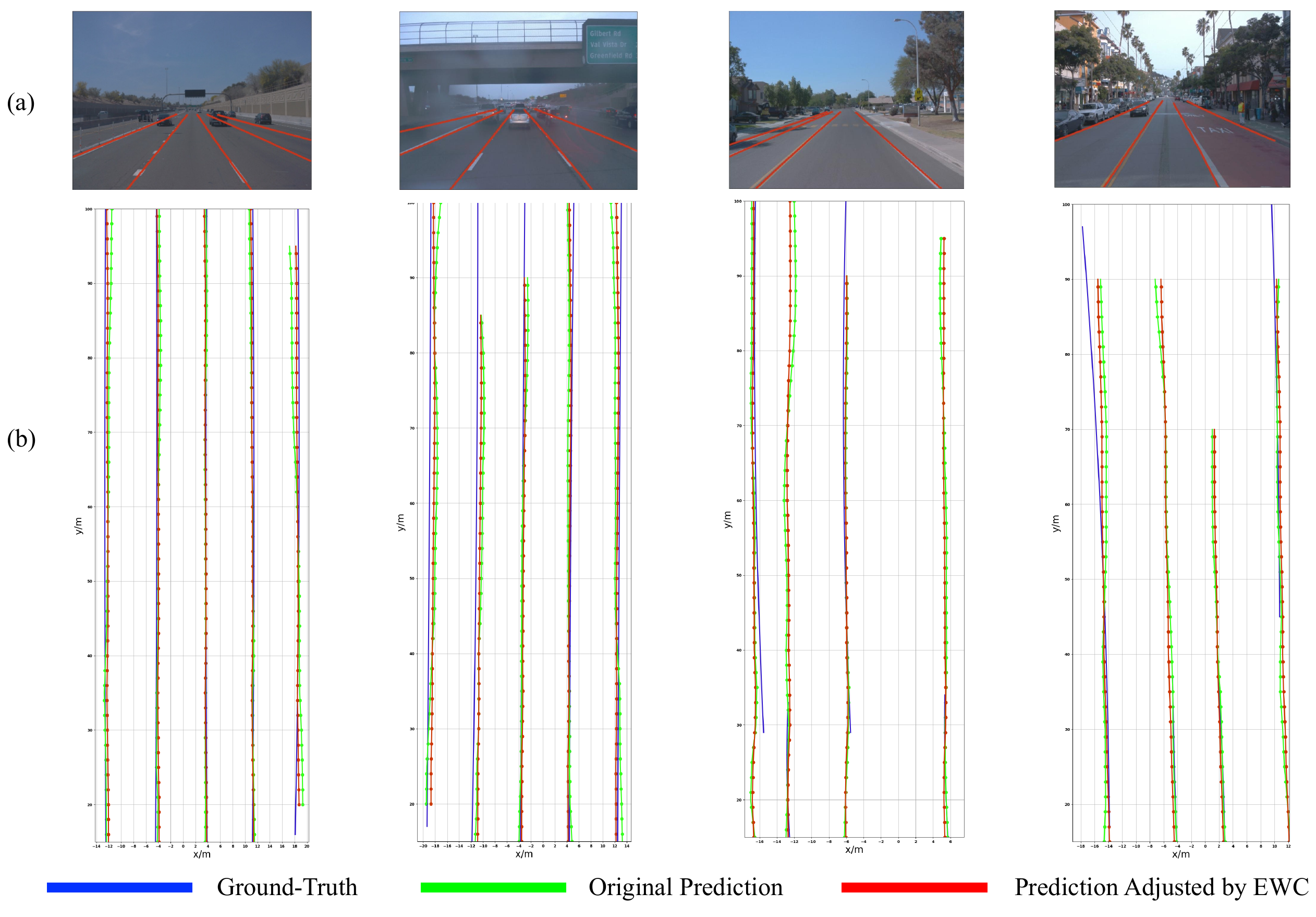}
    \caption{Visualization of equal-width constraint (EWC). (a) Results on 2D images after EWC adjustment. (b) Results on the x-y plane.}
    \label{fig:vis_ewc}
\end{figure*}

\textbf{Equal-Width Constraint.}
We show the visualization results of equal-width constraint (EWC) optimization in Figure~\ref{fig:vis_ewc}.
After adjusting the x coordinates of lanes with EWC, lane predictions are parallel to each other and errors in the distant parts are reduced as well.
It is also worth noting that the ground-truth lanes do not satisfy the equal-width hypothesis in the close parts of some cases, which is possibly due to annotation defects (column $3$).
Therefore, adjusting with EWC may not be beneficial to reducing x error.

\end{appendices}

\end{document}